\documentclass[11pt,a4paper]{article}
\usepackage[hyperref]{acl2020}
\usepackage{times}
\usepackage{latexsym}
\usepackage{multirow}
\usepackage{graphicx}
\usepackage{amsmath}
\usepackage{color}

\usepackage{microtype}

\aclfinalcopy 
\def\aclpaperid{1343} 

\title{Paraphrase Augmented Task-Oriented Dialog Generation}

\author{Silin Gao$^{1}$\thanks{\ \ \ This work was partly done during Silin Gao's summer internship at University of California, Davis.} , Yichi Zhang$^{1}$, Zhijian Ou$^{1,2}$, Zhou Yu$^3$ \\
 $^1$ Speech Processing and Machine Intelligence Lab, Tsinghua University, Beijing, China \\
 $^2$ Beijing National Research Center for Infromation Science and Technology, China \\
 $^3$ University of California, Davis, United States \\
 $^1$ {\tt \{gsl16,zhangyic17\}@mails.tsinghua.edu.cn} \\ $^2$ {\tt ozj@tsinghua.edu.cn}, $^3$ {\tt joyu@ucdavis.edu}}

\begin{document}
\maketitle
\begin{abstract}
Neural generative models have achieved promising performance on dialog generation tasks if given a huge data set. However, the lack of high-quality dialog data and the expensive data annotation process greatly limit their application in real-world settings. We propose a paraphrase augmented response generation (PARG) framework that jointly trains a paraphrase model and a response generation model to improve the dialog generation performance. We also design a method to automatically construct paraphrase training data set based on dialog state and dialog act labels.  PARG is applicable to various dialog generation models, such as TSCP \cite{lei2018sequicity} and DAMD \cite{zhang2019task}.
Experimental results show that the proposed framework improves these state-of-the-art dialog models further on CamRest676 and MultiWOZ. PARG also significantly outperforms other data augmentation methods in dialog generation tasks, especially under low resource settings.
\footnote{This work is supported by NSFC (No.61976122), Ministry of Education and China Mobile joint funding (No.MCM20170301).}
\footnote{The code is available at \url{https://github.com/Silin159/PARG}}

\end{abstract}

\section{Introduction}
Task-oriented dialog systems that are applied to restaurant reservation and ticket booking have attracted extensive attention recently \cite{young2013pomdp,wen2017network,bordes2016learning,eric2017key}. 
Specifically, with the progress on sequence-to-sequence (seq2seq) learning \cite{sutskever2014sequence}, neural generative models have achieved promising performance on dialog response generation \cite{zhao2017generative,lei2018sequicity,zhang2019task}.

However, training such models requires a large amount of high-quality dialog data.
Since each dialog is collected through a human-human or human-machine interaction, it is extremely expensive and time-consuming to create large dialog dataset covering various domains \cite{budzianowski2018multiwoz}. 
After dialogs are collected, we also need to annotate dialog states and dialog acts, which are then used to train language understanding models and learn dialog policy. 
Hiring crowd-sourcing workers to perform these annotations is very costly. 
Therefore, we propose automated data augmentation methods to expand existing well-annotated dialog datasets, and thereby train better dialog systems.


We propose to augment existing dialog data sets through paraphrase. Paraphrase-based data-augmentation methods have been proved to be useful in various tasks, such as machine translation \cite{callison2006improved}, text classification \cite{zhang2015character}, question answering \cite{fader2013paraphrase} and semantic parsing \cite{jia2016data}. All these approaches first find a set of semantically similar sentences. However, finding isolated similar sentences are not enough to construct a dialog utterances' paraphrase. Because an utterance's paraphrase must fit the dialog history as well. For example, when the system says \textit{``Do you prefer a cheap or expensive restaurant?"}, the user may state his intent of asking for a cheap restaurant by \textit{``Cheap please."} or \textit{``Could you find me a cheap restaurant?" }. However, the latter is obviously an improper response which is not coherent with the system question. In other words, a paraphrased dialog utterance needs to serve the same function as the original utterance under the same dialog context. Therefore, we propose to construct dialog paraphrases that consider dialog context in order to improve dialog generation quality.

We also propose the Paraphrase Augmented Response Generation (PARG), an effective learning framework that jointly optimizes dialog paraphrase and dialog response generation. To obtain dialog paraphrases, we first find all the user utterances that serve the same function in different dialogs, such as different ways of asking for Italian food. Then we select the utterances that have the same semantic content but different surface form, to construct a high-quality dialog paraphrase corpus. The corpus is then used to train a paraphrase generation model to generate additional user utterances. Finally, the augmented dialog data is used to train a response generation model. We leverage the multi-stage seq2seq structure \cite{lei2018sequicity,zhang2019task} for both paraphrase and response generation. Moreover, these two models are connected through an additional global attention \cite{bahdanau2014neural} between their decoders, so they can be optimized jointly during training.

In our experiments, we apply our framework on two state-of-the-art models, TSCP \cite{lei2018sequicity} and DAMD \cite{zhang2019task} on two datasets CamRest676 \cite{wen2017network} and MultiWOZ \cite{budzianowski2018multiwoz}, respectively. After applying our framework, the response generation models can generate more informative responses that significantly improves the task completion rate. In particular, our framework is extremely useful under low-resource settings. Our paraphrase augmented model only needs 50\% of data to obtain similar performance of a model without paraphrase augmentation. Our proposed method also outperforms other data augmentation methods, and its comparative advantage increases in settings where only a small amount of training data is available. 

\section{Related Work}

\textbf{Data Augmentation} has been used in various machine learning tasks, such as object detection \cite{redmon2016you} and machine translation \cite{fadaee2017data}. It aims to expand training data to improve model performance. In computer vision, many classical data augmentation methods such as random copy \cite{krizhevsky2012imagenet} and image pair interpolation \cite{zhang2017mixup} have been widely used. 

However, those approaches are not applicable for natural language processing since language is not spatially invariant like images. The word order in a sentence impacts its semantic meaning \cite{zhang2015character}. Therefore, human language augmentation methods aim to generate samples with the same semantic meaning but in different surface forms. Such an idea led to recent augmentation work on the language understanding task \cite{hou2018sequence, kim2019data, yoo2019data, zhao2019data}. However, there is no data augmentation work on task-oriented dialog generation.


\textbf{Paraphrase} is the technique that generates alternative expressions. Most of the existing work on paraphrase aims to improve the quality of generated sentences. For example, phrase dictionary \cite{cao2017joint} and semantic annotations \cite{wang2019task} are used to assist the paraphrase model to improve the language quality. To make a controllable paraphrase model, syntactic information \cite{iyyer2018adversarial} is also adopted. And, recently, different levels of granularity \cite{li2019decomposable} are considered to make paraphrase decomposable and interpretable. In this paper, we utilize a language environment to assist paraphrase, and use paraphrase as a tool to augment the training data of dialog systems.

\section{Proposed Framework}

In this section, we first introduce how to construct a  paraphrase dataset to train paraphrase generation models.
Then we describe the work flow of the proposed PARG model. 

\subsection{Paraphrase Data Construction} \label{pdc}
We propose a three-step procedure to find dialog utterances that are a paraphrase of each other. 
First, we perform delexicalization to pre-process dialog utterances to reduce the surface form language variability.
Then for each user utterance, we match the utterances in other dialogs that play the same function to find its paraphrase candidates. 
Finally, we filter out unqualified paraphrases which have a low semantic similarity or a low surface form diversity comparing to the original utterance. 

Similar to the delexication process introduced in \citet{henderson2014robust}, we replace the slot values in each utterance by their slot name indicator. 
For example, the user utterance \textit{``I want a cheap restaurant."} is delexicalized as \textit{``I want a [pricerange] restaurant."}. 
The slot values can be dropped since their varieties only influence the database search results but have no impact on how the dialog progresses. In other words, no matter whether the user is asking for a cheap or an expansive restaurant, he represents the same intent of requesting a restaurant with a specific price range in the dialog. Therefore through delexicalization, the language variations brought by numerous slot values can be reduced, thus it is easier to find paraphrases. 

\begin{figure}[t]
\centering
\includegraphics[width=1.0\columnwidth]{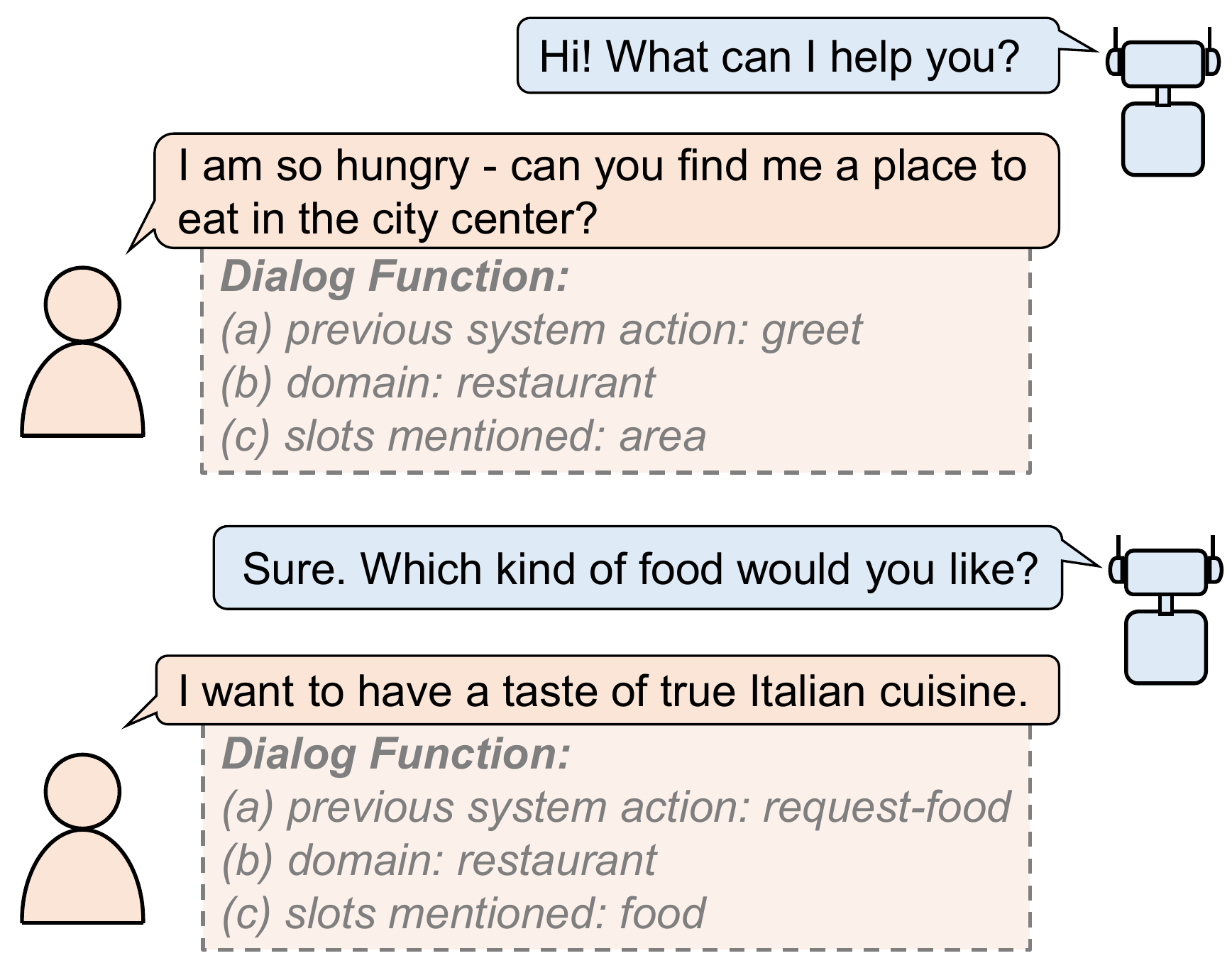}
\caption{Illustration of the dialog function of each turn's user utterance.}
\label{df}
\end{figure}

After delexicalization, we find utterances that play the same role or serve the same \textit{dialog function} in different dialogs. We denote the dialog function of turn $t$ as $DF_t$. It consists of three types of information: 1) current dialog domain $D_t$, 2) slots mentioned $S_t$ in the current turn, and 3) system's dialog act $A_{t-1}$ in the previous turn, which is formulated as: 
\begin{equation}
    DF_{t}=( D_{t}, S_t, A_{t-1})
\end{equation}
The slots mentioned represent the key information towards task completion, which is the most important information to determine the function of the utterance. The dialog domain is included in the function to avoid ambiguities brought by slots that shared across different domains. For example, asking for the location of a hotel is different from asking for a restaurant. The previous system act is considered to ensure a coherent dialog context, since each turn's user utterance is a reply to the previous system response. Fig.\ref{df} gives out an example of dialog function. For each user utterance in the dialog dataset, we go through all the available data and find all utterances with the same dialog function as paraphrase candidates of it. 

As each utterance may have many paraphrase candidates, we only keep the high-quality paraphrase pairs that are similar in semantic but different in surface form. We use the BLEU \cite{papineni2002bleu} score and the diversity score proposed in \citet{hou2018sequence} to evaluate the paraphrase quality. Specifically, if the BLEU score is too low (below 0.2 in our experiments) we consider the paraphrase pair as semantically irrelevant and filtered it out. If the diversity score is too low (below 3.4 in our experiments) we discard the paraphrase pair since it is too alike in terms of surface form language. To find a paraphrase for each of those utterances that do not have any, we gradually reduce the filter threshold of diversity score.

\begin{figure}[t]
\centering
\includegraphics[width=1.0\columnwidth]{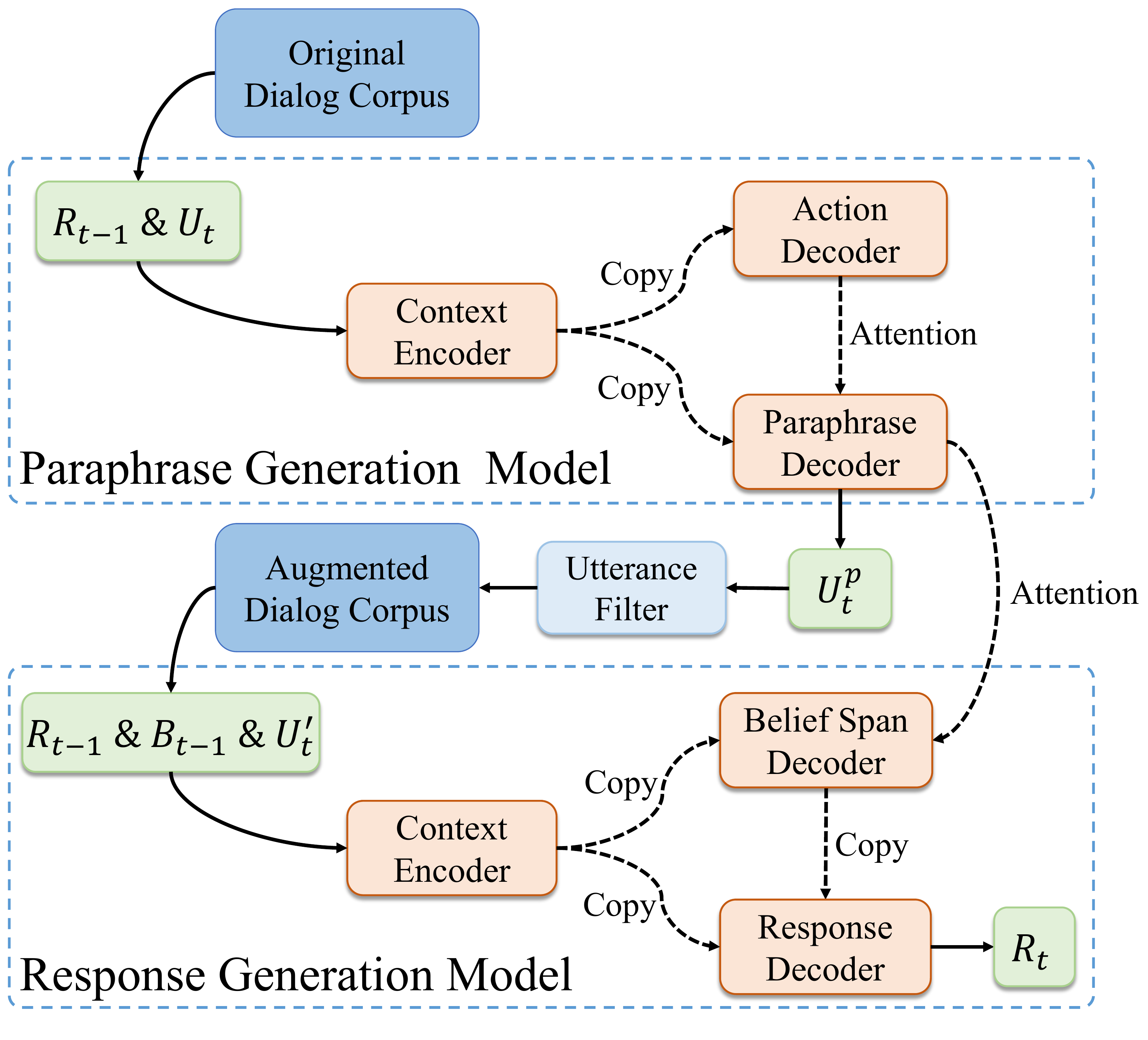}
\caption{Overview of our Paraphrase Augmented Response Generation (PARG) framework. Solid arrows denote the input or output word sequence. Dash arrows denote hidden states shared between modules. $U_{t}$, $B_{t}$ and $R_{t}$ represent turn $t$'s user utterance, dialog state and system response respectively. $U_{t}^{p}$ represents the paraphrase utterance generated by the paraphrase model. The input of the generation model can be either the generated $U_{t}^{p}$ or $U_{t}$, denoted as $U_{t}^{'}$, together with the corresponding dialog state and previous system response.}
\label{framework}
\end{figure}

\subsection{Paraphrase Augmented Response Generation}
Figure \ref{framework} shows an overview of our paraphrase based data augmentation framework. It consists of a paraphrase generation model, a low-quality paraphrase filter and a response generation model. We describe each module in detail below. 

\textbf{Paraphrase Generation Model.} 
Our paraphrase generation model has a seq2seq architecture with a context encoder and two decoders for action decoding and paraphrase decoding. 
The context encoder takes the concatenation of previous system response $R_{t-1}$ and current user utterance $U_t$ as input and encodes them into hidden states. 
Then the hidden states are used to decode the previous system action $A_{t-1}$, where the system action is also a sequence of tokens that first introduced in \citet{zhang2019task}. 
Finally the paraphrase decoder decodes the paraphrase $U_t^p$ based on the hidden states of both the encoder and the action decoder. 
\begin{align}
    h^{A_{t-1}}&=\mathrm{Seq2Seq}(R_{t-1},U_{t}) \\
    U_t^p&=\mathrm{Seq2Seq}(R_{t-1},U_{t}|h^{A_{t-1}})
\end{align}
where $h^{A_{t-1}}$ denotes the hidden states of the action decoder. 
We leverage copy mechanism \cite{gu2016incorporating} to copy words from input utterances to previous system action and paraphrase. 
The action decoding process is used to help paraphrase decoding through an attention connection between the decoders, whose significance lies in improving dialog context awareness. 

\textbf{Paraphrase Filter.} We then send the generated paraphrase into a filter module to determine if it qualifies as an additional training instance. We aim to keep paraphrases that can serve the same dialog function with the original utterance. So we filter out paraphrases that did not include all of the slots mentioned in the original utterance. Besides, we also filter out paraphrases that have a different meaning and/or a similar surface form compared to the original utterance by the same way in our paraphrase data construction process. We still use 0.2 and 3.4 as the thresholds for BLEU and diversity score respectively in our experiments.

\textbf{Response Generation Model.} We use two state-of-the-art seq2seq model, TSCP \cite{lei2018sequicity} and DAMD \cite{zhang2019task} for single domain and multi-domain response generation respectively. We will describe the workflow of our framework based on the TSCP model, as shown in Fig.\ref{framework}. For DAMD the process is similar since the only difference between these two models is that DAMD has an additional action span decoder between the belief span decoder and the response decoder.
The model input is the concatenation of the current user utterance $U_{t}^{'}$, the previous belief span $B_{t-1}$ (slots mentioned by user) and the system response $R_{t-1}$, where $U_{t}^{'}$ is either the original user utterance $U_{t}$ or its paraphrase $U_{t}^{p}$ generated by the paraphrase generation model.
The model is a two-stage decoding network, where the belief span and system response are decoded sequentially using the copy mechanism. Specifically, we introduce an attention connection between the paraphrase decoder and the belief span decoder to allow the gradient in the response generation model to back-propagate to the paraphrase generation model. So the response generation model can guide the paraphrase decoder to generate better paraphrases and vice versa. This process can be formulated as:
\begin{align}
    h^{B_{t}}&=\mathrm{Seq2Seq}(R_{t-1},U_{t}^{'},B_{t-1}|h^{U_{t}^p}) \\
    R_t&=\mathrm{Seq2Seq}(R_{t-1},U_{t}^{'}|h^{B_{t}})
\end{align}
where $h^{B_{t}}$ and $h^{U_{t}^{p}}$ denote the hidden states of the belief span decoder and paraphrase decoder respectively.  


\textbf{Training and Evaluation.} 
The model is joint optimized through supervised learning. Specifically, the system action labels, the paraphrase data (collected through the process introduces in the previous section), the dialog state labels and the reference response are used to calculate the cross-entropy loss of the four decoders, denoted as $loss_{a}$, $loss_{p}$, $loss_{b}$, and $loss_{r}$, respectively. Then we calcualte the sum of all the losses and perform gradient descent for training. The total loss function for training are formulated as:

\begin{equation}
    loss = loss_{a} + loss_{p} + loss_{b} + loss_{r}
\end{equation}

Note that we only augment user utterance as additional input utterances during training. We alternatively use the original $U_t$ and generated $U_t^p$ as input to the response generation model, while other elements such as belief spans and responses remain the same. Since both decoders are forced to recognize more user expressions, the language understanding and response generation performance improve simultaneously.  
If the generated $U_t^p$ is in low quality and filtered out, only the original $U_t$ is used to train the response generation model in that turn. This often happens at the beginning of training when the paraphrase model is under-fitting. 
During testing, only the ground truth user utterances are used as input. However, we still utilize the paraphrase generation model to compute attention between the paraphrase decoder and the belief span decoder. This is because we believe that the paraphrase decoding process can help the belief span decoding process since it provides additional explanations of the user utterance.

\section{Experimental Settings}

\subsection{Datasets and Evaluation Metrics}
We conduct our experiments based on two datasets, CamRest676 \cite{wen2017network} and MultiWOZ \cite{budzianowski2018multiwoz}.
Dialogs in both are collected through crowd-sourcing on the Amazon Mechanical Turk platform.
Besides experiments on the full datasets, we also conduct experiments using only 20\% or 50\% of dialog data for training to evaluate the promotion through data augmentation under low-resource settings. 

\textbf{CamRest676} is a single domain dataset consisting of dialogs about restaurant reservation. The dataset has 676 dialogs which are split into training, development and testing set by the ratio of 3:1:1. The average number of turns is 4.06. There are 3 slot types and 99 allowable values in the task ontology. 
We use three metrics for evaluation following \citet{lei2018sequicity}. \textbf{Entity Match Rate} (EMR) is the proportion that the system capture the correct user goal. \textbf{Success F1} (Succ.F1) score measures whether the system can provide correct information requested by user. While these two metrics are used for evaluating system's task completion ability, we use \textbf{BLEU} \cite{papineni2002bleu} to evaluate the language fluency of generated responses.

\textbf{MultiWOZ} is a challenging large-scale multi-domain dataset proposed recently \cite{budzianowski2018multiwoz}. It consists of dialogs between tourists and clerks at an information center, across seven domains including hotel, restaurant, train, etc. There are 8433/1000/1000 dialogs in training, development and testing set respectively, and the number of turn is 6.85 on average. Meanwhile, MultiWOZ has a complex ontology with 32 slot types and 2,426 corresponding slot values. We use the evaluation metrics proposed by \citet{budzianowski2018multiwoz}, which are how often the system provides an correct entity (\textbf{inform rate}) and answers all the requested information (\textbf{success rate}), and how fluent the response is (\textbf{BLEU}). We also report a combined score computed via $(Inform+Success)\times0.5+BLEU$ for overall quality measure as suggested in \cite{mehri2019structured}.

\begin{table*}[t]
\centering
\resizebox{0.8\textwidth}{!}{
\begin{tabular}{c|ccc|ccc|ccc}
\hline
\multirow{2}{*}{Model} & \multicolumn{3}{|c|}{20\% Data} & \multicolumn{3}{|c|}{50\% Data} & \multicolumn{3}{|c}{Full Data}\\
\cline{2-10}
 & BLEU & EMR & Succ.F1 & BLEU & EMR & Succ.F1 & BLEU & EMR & Succ.F1 \\
\hline
TSCP & 0.154 & 0.791 & 0.806 & 0.225 & 0.853 & 0.817 & 0.253 & 0.927 & 0.854 \\
\hline
WordSub & 0.140 & 0.821 & 0.818 & 0.212 & 0.866 & 0.822 & 0.239 & 0.930 & 0.846 \\
TextSub & 0.144 & 0.834 & 0.826 & 0.220 & 0.895 & 0.831 & 0.245 & 0.942 & 0.850 \\
UtterSub & 0.149 & 0.826 & 0.829 & 0.216 & 0.876 & 0.838 & 0.245 & 0.938 & 0.852 \\
\hline
NAEPara & \textbf{0.155} & 0.830 & 0.831 & 0.222 & 0.891 & 0.843 & 0.251 & 0.940 & 0.855 \\
SRPara & 0.154 & 0.832 & 0.826 & \textbf{0.228} & 0.886 & 0.840 & \textbf{0.254} & 0.938 & 0.852 \\
\hline
PARG &\textbf{0.155} & \textbf{0.852} & \textbf{0.849} & 0.226 & \textbf{0.908} & \textbf{0.853} & 0.252 & \textbf{0.943} & \textbf{0.861} \\
\hline
\end{tabular}
}
\caption{Results on CamRest676. The best scores are in bold. }
\label{table1}
\end{table*}

\begin{table*}[t]
\centering
\resizebox{0.95\textwidth}{!}{
\begin{tabular}{c|cccc|cccc|cccc}
\hline
\multirow{2}{*}{Model} & \multicolumn{4}{|c|}{20\% Data} & \multicolumn{4}{|c|}{50\% Data} & \multicolumn{4}{|c}{Full Data}\\
\cline{2-13}
 & BLEU & Info & Succ & Comb & BLEU & Info & Succ & Comb & BLEU & Info & Succ & Comb\\
\hline
DAMD & 0.121 & 0.779 & 0.703 & 0.862 & 0.169 & 0.830 & 0.729 & 0.948 & 0.183 & 0.895 & 0.758 & 1.009 \\
\hline
WordSub & 0.119 & 0.783 & 0.712 & 0.866 & 0.166 & 0.821 & 0.736 & 0.944 & 0.176 & 0.882 & 0.754 & 0.994 \\
TextSub & 0.123 & 0.813 & 0.719 & 0.889 & 0.174 & 0.841 & 0.741 & 0.965 & 0.182 & 0.890 & 0.760 & 1.007 \\
UtterSub & 0.112 & 0.802 & 0.714 & 0.870 & 0.169 & 0.853 & 0.737 & 0.964 & 0.179 & 0.893 & 0.761 & 1.006 \\
\hline
NAEPara & 0.126 & 0.820 & 0.723 & 0.898 & 0.164 & 0.850 & 0.750 & 0.964 & 0.179 & 0.893 & 0.761 & 1.006 \\
SRPara & \textbf{0.130} & 0.817 & 0.725 & 0.901 & \textbf{0.175} & 0.864 & 0.753 & 0.984 & 0.186 & 0.903 & 0.773 & 1.024 \\
\hline
PARG & 0.127 & \textbf{0.825} & \textbf{0.739} & \textbf{0.909} & 0.172 & \textbf{0.878} & \textbf{0.768} & \textbf{0.995} & \textbf{0.188} & \textbf{0.911} & \textbf{0.789} & \textbf{1.038} \\
\hline
\end{tabular}
}
\caption{Results on MultiWOZ. The best scores are in bold.}
\label{table2}
\end{table*}

\subsection{Implementation Settings}
We use a one-layer, bi-directional GRU as the context encoder and two standard GRU as the action decoder and paraphrase decoder. The embedding size and hidden size are both 50 on CamRest676 and 100 for MultiWOZ. The copy mechanism and attention connection are added as shown in Fig.\ref{framework}. For the response generation model, we leverage the state-of-the-art model on each dataset, which is the Two-stage Copy Net (TSCP) \cite{lei2018sequicity} for CamRest676 and Domain Aware Multi-Decoder (DAMD) \cite{zhang2019task} for MultiWOZ. 
We use the model structures that follow the default settings in the open source implementation of TSCP\footnote{https://github.com/WING-NUS/sequicity} and DAMD\footnote{https://gitlab.com/ucdavisnlp/damd-multiwoz}. We use the the Adam optimizer \cite{kingma2014adam} with a learning rate of 0.003 and 0.005 for CamRest676 and MultiWOZ, respectively. We halve the learning rate when the total loss of our model on development set does not reduce in three consecutive epochs, and we stop the training when the total loss does not reduce in five consecutive epochs.  We set the learning rate to 0.0001 and the decay parameter to 0.8 during reinforcement fine tuning in TSCP.


\subsection{Baseline Methods}
We compare the proposed method with five other data augmentation methods, three of which are based on text replacement and the other two are based on neural paraphrase generation models.


\begin{itemize}
    \item \textbf{WordSub} denotes the rare word substitution method proposed by  \citet{fadaee2017data}. It generates new sentences by replacing common words with rare ones. A bi-directional LSTM language model is trained to select the proper substitution words. We do not substitute key words associated with slot values to maintain the dialog function of utterances.
    
    \item \textbf{TextSub} denotes the text span replacement method proposed by \citet{yin2019dialog}. It replaces a sequence of tokens (text span) by their paraphrase candidates from the lexicon database (PPDB \cite{pavlick2015ppdb}). The selection of text spans is based on a policy network, which is trained jointly with the belief span decoder through reinforcement learning. The slot values are also fixed with the same purpose as in WordSub. 
    
    \item \textbf{UtterSub} denotes the simple utterance replacement augmentation. We use the paraphrases obtained in dialog dataset as new training samples directly instead of training the paraphrase model to generate new samples.
    
    \item \textbf{NAEPara} denotes a paraphrase model with single encoder-decoder structure. This model, denoted as noising auto-encoder (NAE) in \citet{li2019insufficient}, injects random noise to the encoder's hidden states to improve generation varieties, which has proven to be effective in \cite{kurata2016labeled}. For model implementation, we use the same GRU nets as in our paraphrase model. And we multiply perturbations, sampled from the uniform distribution between 0.6 and 1.4, to the encoder's hidden states when generating paraphrases.
    
    \item \textbf{SRPara} denotes a paraphrase model with SR-PB \cite{wang2019task} structure. In this structure, a semantic parser SLING \cite{ringgaard2017sling} is used to analyze the semantic frame of an utterance and the semantic role of each token in it. Then the sequences of token, semantic frame labels and semantic role labels are fed into three parallel encoders separately. The outputs of the three encoders are projected through a linear layer, and then sent to a decoder to generate the paraphrase. The implementation of encoders and the decoder is the same as NAEPara.
    
\end{itemize}

We utilize the same dataset (CamRest676 or MultiWOZ) to train all the models for fair comparison. Specifically, we use all the user utterances in the training corpus of CamRest676 or MultiWOZ to train the LSTM language model of WordSub and the policy network of TextSub. And we use the same paraphrase data constructed in \ref{pdc} to train the paraphrase models in NAEPara and SRPara.



\section{Results and Analysis}
The experimental results on CamRest676 and MultiWOZ are shown in Table \ref{table1} and Table \ref{table2}, respectively. In both tables, the first line is the baseline results without data augmentation, the second to sixth lines are results obtained by different data augmentation methods (substitution-based or paraphrase-based), and the last line is the performance of our proposal. The results are grouped into three columns according to the size of training data (20\%/50\%/full). 

We observe some common conclusions supported by the experimental results on both datasets. First, our proposed data augmentation framework significantly improves the system's task completion ability (EMR, Succ.F1, Info and Succ) consistently without harming the language fluency. This indicates that incorporating additional dialog paraphrases is beneficial for learning informative responses, since more user expressions are seen by the model during training. 

Secondly, our framework outperforms other data augmentation methods in terms of dialog task relevant metrics under all circumstances. 
In particular, paraphrase based methods are more likely to produce more fluent and informative responses than local substitution methods (WordSub and TextSub), because neural generative models consider dialog history to generate more coherent utterances. 
The improvement of PARG over UtterRep suggests that our paraphrase generation model provides a more robust way of utilizing the additional information contained in paraphrases. 
Our paraphrase generation model outperforms other paraphrase based methods (NAEPara and SRPara) since the decoding process of previous system action and the gradient back-propagation through the belief span decoder provide strong dialog context information for paraphrase generation. 

Thirdly, the less data is available, the more improvement can be achieved through our data augmentation. It is worth noting that after applying PARG, the model trained on only 50\% data obtain comparable results to the model trained on the full dataset without data augmentation, in terms of task relevant metrics. The similar results are also observed by comparing the models trained on 20\% data with augmentation and 50\% data without augmentation. This indicates that our method is of great significance under low resource settings. 

PARG sometimes gets a slightly lower BLEU score compared to other methods. This is potentially because that although seq2seq models can learn responses which corresponding to a correct action, the surface language can still vary among training and testing utterances due to the natural variety of human languages. Therefore, the BLEU score, which measures the likeness of surface language, may drop despite the system generate good functional responses. 

We also observe some diverse results on CamRest676 and MultiWOZ. Under the full data setting, the improvement gained by our data augmentation method on CamRest676 is lower than on MultiWOZ, since the single domain task in CamRest676 is easy and the data is enough for model training without conducting augmentation. While for MultiWOZ, due to large language variations and the complex ontology, the utterance space is not well-explored, thus the response generation process can benefit more through incorporating additional dialog data.  


\section{Ablation Study}
In this section we investigate the function of each component in our paraphrase augmented response generation framework. In particular, we discard 1) the act decoder (PARG w/o Act), 2) the utterance filter (PARG w/o Filt) or 3) joint training (PARG w/o Join) one at a time, then do model training and evaluation on the full MultiWOZ dataset. The results are shown in Table \ref{table3}. 

\begin{table}[h]
\centering
\resizebox{1.0\columnwidth}{!}{
\smallskip\begin{tabular}{c|cccc}
\hline
Model & BLEU & Info & Succ & Comb \\
\hline
DAMD & 0.183 & 0.895 & 0.758 & 1.009 \\
\hline
PARG & 0.188 & 0.911 & 0.789 & 1.038 \\
\hline
PARG w/o Filt & 0.173 & 0.887 & 0.765 & 0.999 (-0.039) \\
PARG w/o Act & 0.180 & 0.897 & 0.763 & 1.010 (-0.028) \\
PARG w/o Join & 0.185 & 0.905 & 0.782 & 1.028 (-0.010) \\
\hline
\end{tabular}
}
\caption{Ablation results on MultiWOZ. The changes of combine score compared to PARG are shown in parentheses. }
\label{table3}
\end{table}

\begin{table*}[t]
\centering
\resizebox{1.0\textwidth}{!}{
\begin{tabular}{c|c|l}
\hline
\multicolumn{3}{l}{User Utterance: Can you help me find a restaurant in the south that doesn't cost a lot of money.} \\
\multicolumn{3}{l}{Ground Truth Dialog State: pricerange=cheap, area=south} \\
\multicolumn{3}{l}{Reference Response: Nandos is a nice place, it serves Portuguese food. Is there anything else?} \\
\hline
\multirow{4}*{Full Data} & \multirow{2}*{TSCP} &  Generated Dialog State: pricerange=cheap, area=south \\
 & & Generated Response: Nandos is a restaurant in the south. Would you like something different? \\
 \cline{2-3}
 & \multirow{2}*{PARG} & Generated Dialog State:  pricerange=cheap, area=south \\
 & &  Generated Response: Nandos is a Portuguese restaurant in the south. Anything else you need?\\
\hline
\multirow{4}*{50\% Data} & \multirow{2}*{TSCP} &  Generated Dialog State: area=south \\
 & & Generated Response: Taj Tandoori is an Indian restaurant, it is in the expensive price range.\\
 \cline{2-3}
 & \multirow{2}*{PARG} & Generated Dialog State:  pricerange=cheap, area=south \\
 & &  Generated Response: Nandos serves Portuguese food. Would you like the address?\\
\hline
\end{tabular}
}
\caption{Comparison of response generation results before and after applying our paraphrase augmented method. Models trained on full data and 50\% data are compared respectively.}
\label{table4}
\end{table*}

\begin{table*}[t]
\centering
\resizebox{1.0\textwidth}{!}{
\begin{tabular}{l|l}
\hline
\multicolumn{1}{c|}{Dialog Function} & \multicolumn{1}{|c}{Utterance Paraphrase}  \\
\hline
Domain: train & Previous Response: What time would you like to leave from norwich?\\
Slots Mentioned: leave & Original Utterance: I would like to leave at 14:45. What is the price? \\
Previous System Act: request-leave & Matched Paraphrase: 14:45, please. What is the duration of the train ride? \\
\hline
Domain: hotel & Previous Response: Acorn Guest House is available if that works for you. \\
Slots Mentioned: parking & Original Utterance: That is good. And I need a free parking, does it have? \\
Previous System Act: inform-name & Matched Paraphrase: This place is fine. Is it near a hotel with free parking?\\
\hline
\end{tabular}
}
\caption{Examples of ill-matched paraphrase pairs obtained by our paraphrase matching method. }
\label{table5}
\end{table*}

\begin{table}[t]
\centering
\resizebox{0.95\columnwidth}{!}{
\smallskip\begin{tabular}{c|c}
\hline
\multicolumn{2}{l}{Original Utterance:} \\
\multicolumn{2}{l}{I need an inexpensive restaurant on the north side.} \\
\hline
\multirow{2}*{TextSub} & I'm looking for place inexpensive \\
& restaurant is located in the north.\\
\hline
\multirow{2}*{SRPara} & Please find me an inexpensive \\
& restaurant in the north part of the town.\\
\hline
\multirow{2}*{PARG} & Can you recommend me a cheap \\
& restaurant in the north area.\\
\hline
\end{tabular}
}
\caption{Paraphrased utterances generated by different methods. }
\label{table6}
\end{table}

We observe that removing the utterance filter brings the biggest drop in response quality in terms of combined score (-0.039). This suggests the importance of using only high-quality paraphrases to train the response generation model, because the ill generation utterances will introduce errors to the downstream model. The model also suffers from a performance drop (-0.028) after removing the previous system action decoder, which indicates that the supervision from previous system action labels is beneficial for generating better paraphrases. Finally, we train the paraphrase generation model and reponse generation model separately and oberve a slight drop of combined score (-0.010). This is because through the attention connection between the paraphrase decoder and belief span decoder, the loss computed for response generation can also guide the paraphrase generation model to generate paraphrases that directly benefit to the response generation process. Although the improvement is relatively marginal, joint training has additional advantages in simplifying the training process. Specifically, we only need to conduct a single run of training and optimize a single set of hyperparameters.

\section{Case Study and Error Analysis}
We conduct several case studies to illustrate the response generation quality, paraphrase generation quality, as well as errors made by our model.

Table \ref{table4} compares the dialog state and system response generated by the original model TSCP to those generated by PARG. We investigate the results from both the 50\% and the full scale CamRest676 experiments, to further show our framework's superiority in low resource scenarios.
On full training data, TSCP and PARG both generate correct dialog state slots. However, TSCP generates a wrong question ``Would you like something different?'', as if no restaurant satisfies the user's request. While PARG generates an appropriate question ``Anything else you need?'' to ask user for further request about the recommended restaurant. 
When we reduce the training data to half, TSCP generates wrong dialog state slots, and therefore recommends an expensive restaurant. But PARG does not suffer from this problem and generates a correct response. This example suggests that PARG can effectively improve the quality of dialog generation in low resource settings.

Although our paraphrase augmented data augmentation framework shows a notable superiority on the dialog generation quality, it still has some limitations. Table \ref{table5} shows some errors that PARG made in our paraphrase data construction process.
In the first case, the question ``What is the price?'' raised by the original utterance doesn't match the question ``What is the duration of the train ride?'' in the paraphrase. This error is made since we do not have user act labels in the dialog datasets. Defining the dialog function of user utterance more precisely by adding its user act can solve this problem.
Another incoherence of paraphrase sources from the switch of dialog domains in multi-domain dialogs. In the example, the word ``place'' in the paraphrase refers to another site irrelevant to the hotel in the previous system response, which might be an attraction or a restaurant. The domain of the previous turn should also be considered in the dialog function to provide more domain information, which is regarded as a potential solution for this issue. 

We also compare the utterances generated by different data augmentation methods to show the superiority of PARG in terms of paraphrase generation quality. We select TextSub and SRPara for comparison, since they are the best replacement-based and paraphrase-based methods achieving the highest combined scores on MultiWOZ respectively.   Table \ref{table6} shows an example of paraphrases generated by the three methods. We find that the paraphrase generated by TextSub is of bad quality because it is not in accordance with normal grammar, while the paraphrase generated by SRPara is fluent and semantically similar to the original utterance. However, the paraphrase generated by our proposed PARG has higher quality. It flexibly changes the rare word ``inexpensive'' to the common word ``cheap'', which enlarges the surface form diversity. The high-quality paraphrases can give better guidance to the downstream response generation model, which explains the significant improvement in terms of task completion rate obtained by PARG.

\section{Human Evaluation}
We conduct human evaluation to further illustrate PARG's superiority in terms of paraphrase generation. We use one-to-one comparison to evaluate the relative quality of paraphrases generated by PARG versus strong
baselines (NAEPara and SRPara).

In our experiments, we advise the judges to evaluate the quality of a paraphrase according to its similarity of user intent with the original utterance. We sample one hundred dialog turns. And in each turn, the paraphrase generated by PARG is given one-to-one comparisons with each baseline's paraphrase by five judges. Specifically, we ask the judges to choose whether the paraphrase generated by PARG is of better, equal or worse quality than the paraphrase generated by NAEPara or SRPara, given the original utterance.

\begin{table}[h]
	\centering
	\resizebox{1.0\columnwidth}{!}{
		\smallskip\begin{tabular}{l|ccc}
			\hline
			Comparison & Better\% & Equal\% & Worse\%\\
			\hline
			PARG vs. NAEPara & 59.2\% & 18.4\% & 22.4\% \\
			PARG vs. SRPara & 55.4\% & 20.8\% & 23.8\% \\
			\hline
		\end{tabular}
	}
	\caption{Human evaluation results.}
	\label{table7}
\end{table}

The results are shown in Table \ref{table7}. We report the percentage of different choices made by the judges in each one-to-one comparison, including the percentage of cases that PARG generates better (Better\%), so-so (Equal\%), or worse (Worse\%) paraphrases. We observe that PARG generates better paraphrases in a large proportion of cases, no matter compared to NAEPara or SRPara. This suggests that PARG outperforms both NAEPara and SRPara in terms of paraphrase generation quality, which further proves that the dialog data augmented by PARG can provide better guidance to the response generation tasks.

\section{Conclusion}
In this paper, we propose to use dialog paraphrase as data augmentation to improve the response generation quality of task-oriented dialog systems. We give out the definition of the paraphrase for a dialog utterance and design an approach to construct paraphrase dataset from a dialog corpus. We propose a Paraphrase Augmented Response Generation (PARG) framework which consists of a paraphrase generation model, an utterance filter and a response generation model, where the models are trained jointly to take fully advantage of the paraphrase data for better response generation performance. Our framework achieves significant improvements when it is applied to state-of-the-art response generation models on two datasets. It also beats other data augmentation methods, especially under the low-resource settings.


\bibliography{main}
\bibliographystyle{acl_natbib}

\end{document}